\documentclass[letterpaper, 10 pt, conference]{ieeeconf}  

\IEEEoverridecommandlockouts                              

\usepackage{graphics} 
\usepackage{epsfig} 
\usepackage{mathptmx} 
\usepackage{times} 
\usepackage{caption}
\usepackage[T1]{fontenc} 
\usepackage{cite}
\usepackage{booktabs}
\usepackage{amsmath,amssymb,amsfonts}
\usepackage{algorithmic}
\usepackage{setspace}
\usepackage{subfigure}
\usepackage{graphicx}
\usepackage{textcomp}
\usepackage[table]{xcolor}
\usepackage{multirow}
\usepackage{threeparttable}
\usepackage{tabularx}
\usepackage{mathtools}
\usepackage[free-standing-units=true]{siunitx}
\usepackage{float}
\usepackage{csquotes}
\newcommand{\rom}[1]{\uppercase\expandafter{\romannumeral #1\relax}}

\usepackage{array}

\newcommand{\PreserveBackslash}[1]{\let\temp=\\#1\let\\=\temp}
\newcolumntype{C}[1]{>{\PreserveBackslash\centering}p{#1}}
\newcolumntype{R}[1]{>{\PreserveBackslash\raggedleft}p{#1}}
\newcolumntype{L}[1]{>{\PreserveBackslash\raggedright}p{#1}}
\definecolor{ourgray}{gray}{0.9}

\makeatletter
\let\NAT@parse\undefined
\makeatother

\usepackage[colorlinks]{hyperref}
  \hypersetup{
    citecolor=blue,
    linkcolor=blue,   
    urlcolor=blue}

\author{Yufeng Jin$^{1,2}$, Niklas Funk$^{1}$, Vignesh Prasad$^{1}$, Zechu Li$^{1}$, Mathias Franzius$^{2}$, Jan Peters$^{1,3,4}$, Georgia Chalvatzaki$^{1,4}$
\thanks{Corresponding author: Yufeng Jin (yufeng.jin@tu-darmstadt.de).}
\thanks{This work was supported by the German Research Foundation (DFG) Emmy Noether Programme (CH 2676/1-1), the EU’s Horizon Europe project \enquote{ARISE} (Grant no.: 101135959), the German Federal Ministry of Education and Research (BMBF) Project \enquote{RiG} (Grant no.: 16ME1001)}
\thanks{$^{1}$Department of Computer Science, TU Darmstadt, Germany.}
\thanks{$^{2}$Honda Research Institute Europe GmbH, Offenbach, Germany.}
\thanks{$^{3}$DFKI, Research Department SAIROL, Darmstadt, Germany.}
\thanks{$^{4}$Hessian.AI, Darmstadt, Germany}
\thanks{This work has been submitted to the IEEE for possible publication. 
Copyright may be transferred without notice, after which this version may no longer be accessible.}
}
\begin{document}

\title{\LARGE \bf SE(3)-PoseFlow: Estimating 6D Pose Distributions\\ for Uncertainty-Aware Robotic Manipulation
}


\maketitle

\begin{abstract}
Object pose estimation is a fundamental problem in robotics and computer vision, yet it remains challenging due to \textit{partial observability}, \textit{occlusions}, and \textit{object symmetries}, which inevitably lead to \textit{pose ambiguity} and multiple hypotheses consistent with the same observation. While deterministic deep networks achieve impressive performance under well-constrained conditions, they are often overconfident and fail to capture the multi-modality of the underlying pose distribution. To address these challenges, we propose a novel probabilistic framework that leverages \textit{flow matching on the SE(3) manifold} for estimating 6D object pose distributions. Unlike existing methods that regress a single deterministic output, our approach models the full pose distribution with a sample-based estimate and enables reasoning about \textit{uncertainty} in ambiguous cases such as symmetric objects or severe occlusions. We achieve state-of-the-art results on Real275, YCB-V and LM-O,
and demonstrate how our sample-based pose estimates can be leveraged in downstream robotic manipulation tasks such as active perception for disambiguating uncertain viewpoints, or guiding grasp synthesis in an uncertainty-aware manner.
\end{abstract}

\begin{keywords}
  Object Pose Uncertainty Estimation, Flow Matching, SE(3) Manifold
\end{keywords}

\section{Introduction}

Estimating the 6D pose of objects is a fundamental problem in robotics, as it enables embodied agents to perceive, manipulate, and interact safely with their environment. In practical applications such as robotic grasping, assembly, and human–robot collaboration, it is not sufficient to output a single deterministic pose estimate. Instead, reasoning about \emph{uncertainty} is critical for ensuring safe and reliable manipulation~\cite{lee2023uncertainpose,armour2023}. Probabilistic models that capture the multi-modality of pose distributions provide richer information than deterministic pose estimates, especially in safety-critical manipulation scenarios where downstream decisions rely on calibrated confidence.

A central challenge in 6D object pose estimation arises from \emph{pose ambiguity}. Symmetries in object geometry, partial observability, and occlusions often yield multiple feasible poses that are indistinguishable from sensor observations~\cite{hsiaoConfrontingAmbiguity6D2024,brachmannUncertaintyDriven6DPose2016,manhardtExplainingAmbiguityObject2019,ikeda2024diffusionnocs,zhangGenPoseGenerativeCategorylevel2023}. Deterministic deep learning approaches such as FoundationPose~\cite{wenFoundationPoseUnified6D2024} have recently advanced the state of the art by leveraging large-scale synthetic training and transformer-based architectures, but they remain limited in their ability to represent multi-hypothesis pose distributions. As a result, they can be over-confident in ambiguous cases, which is undesirable for robotic planning and control. 

\begin{figure}[t]
    \captionsetup{font=small}
    \centering
    \includegraphics[width=\linewidth]{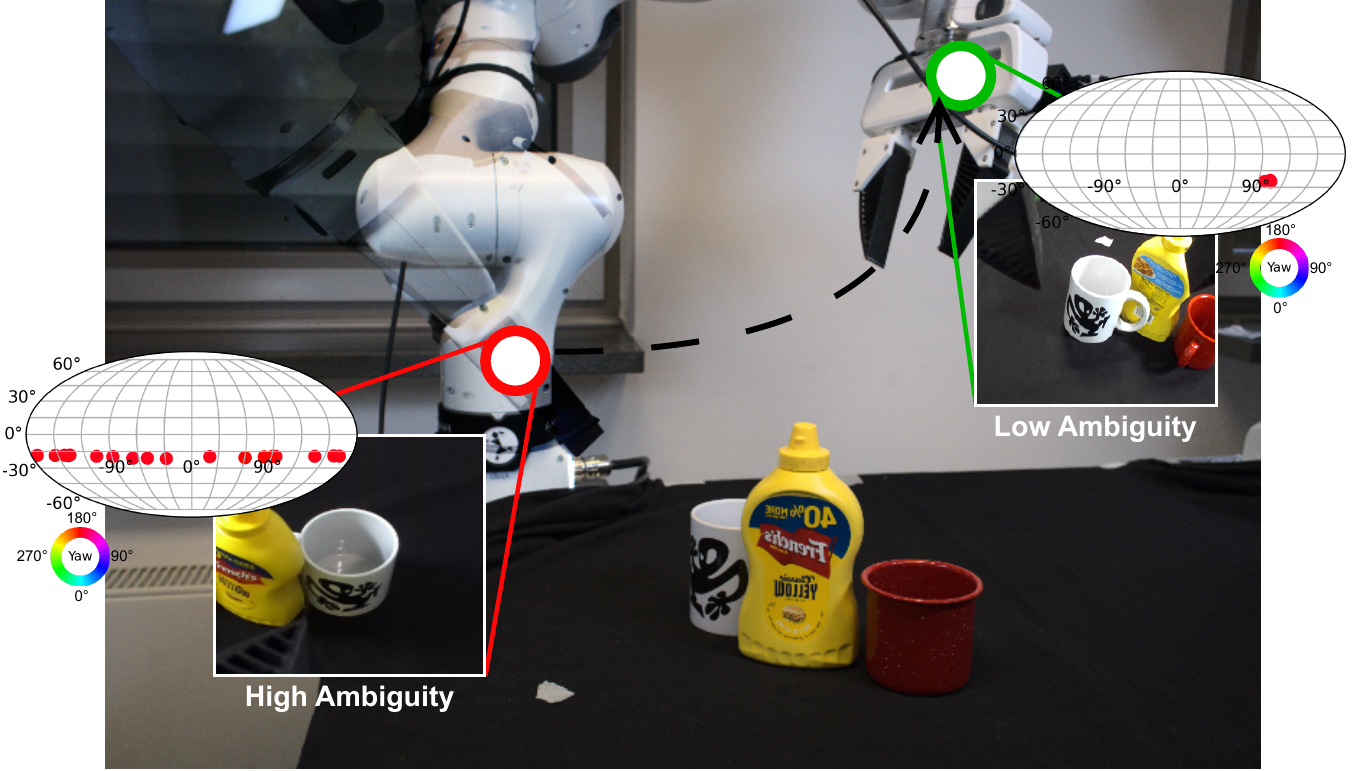}
    \captionsetup{width=\linewidth}
    \caption{\small We propose an uncertainty-aware 6D object pose estimation approach based on $\mathrm{SE}(3)$ flow matching. Our probabilistic framework predicts full 6D pose distributions to handle ambiguities, enabling reliable robotic manipulation under challenging real-world conditions (partial observability, occlusions, and symmetries). SO(3) distributions are visualized on a Mollweide projection, where latitude (pitch) and longitude (roll) map the orientation, and color encodes yaw.}

    \label{fig:active_perception}
    \vspace{-10pt}
\end{figure}

To address these limitations, recent research has explored \emph{probabilistic formulations of object pose estimation}. These methods aim to model the full distribution of feasible poses rather than commit to a single prediction. Early approaches leverage \emph{directional probability distributions} such as the von Mises--Fisher or Bingham distribution to represent rotational uncertainty~\cite{prokudin2018deep,deng2022deep,gilitschenskiDEEPORIENTATIONUNCERTAINTY2020}. While theoretically principled, such parametric models are typically unimodal and require mixtures to capture multi-modal ambiguities, leading to computational inefficiency and numerical instability. More recently, \emph{generative models} based on diffusion~\cite{ikeda2024diffusionnocs,zhangGenPoseGenerativeCategorylevel2023,hsiaoConfrontingAmbiguity6D2024} and normalizing flows~\cite{liuDelvingDiscreteNormalizing2023} have been proposed to directly model complex distributions on $\mathrm{SE}(3)$ through sample-based estimates. These approaches naturally capture multi-modality and uncertainty, but often rely on intermediate representations or remain constrained to synthetic benchmarks. This motivates the development of new methods that combine the scalability of modern architectures with principled probabilistic modeling on the $\mathrm{SE}(3)$ manifold for safe and robust robotics. Our main contributions are summarized as follows:

\begin{itemize}
    \item We propose a probabilistic framework based on \textit{flow matching on the $\mathrm{SE}(3)$ manifold} for 6D object pose estimation. Our method provides a sample-based estimate of the pose distribution that naturally captures uncertainty in ambiguous cases such as object symmetries or severe occlusions.
    \item We introduce an adapted DiT module with masked cross-attention into the $\mathrm{SE}(3)$ flow model, which improves robustness under occlusions and cluttered real-world scenes, thereby achieving competitive state-of-the-art performance across multiple benchmarks.
    \item We demonstrate how the learned $\mathrm{SE}(3)$ distribution can be leveraged for downstream robotic tasks, such as guiding active perception to resolve viewpoint ambiguity, and 
    enabling reliable and effective single-view grasp generation under partial observability.
\end{itemize}

\section{Related Works}
\subsection{Probabilistic Object Pose Estimation}
Recent work has sought to overcome the limits of deterministic regressors by explicitly modeling uncertainty in object pose estimation. One approach leverages \emph{directional probability distributions} for rotational uncertainty: the von Mises--Fisher distribution for Euler angles~\cite{prokudin2018deep}, and the Bingham distribution for unit quaternions~\cite{deng2022deep,gilitschenskiDEEPORIENTATIONUNCERTAINTY2020,okornLearningOrientationDistributions2020,gloverMonteCarloPose,okorn2020learning}. These models handle symmetries well but have drawbacks: (i) computing normalization constants on non-Euclidean manifolds is costly, (ii) they are unimodal and require mixtures for multi-modality, risking mode collapse, and (iii) parameterizations can be unstable and scale poorly.  

Beyond closed-form distributions, \emph{generative probabilistic models} better capture complex, multi-modal pose distributions. DiffusionNOCS~\cite{ikeda2024diffusionnocs} uses image-to-image diffusion to predict NOCS maps aligned with depth, naturally handling symmetry but incurring inference overhead. GenPose~\cite{zhangGenPoseGenerativeCategorylevel2023} applies score-based diffusion on point clouds, sampling multiple hypotheses but perturbing $\mathrm{SO}(3)$ with Gaussian noise and requiring an auxiliary energy network for likelihood estimation. Möller et al.~\cite{moller2024particle} adopt a particle-based diffusion formulation for point clouds, which discards texture cues important for fine-grained alignment. More theoretically, Hsiao et al.~\cite{hsiaoConfrontingAmbiguity6D2024} and Liu et al.~\cite{liuDelvingDiscreteNormalizing2023} study diffusion and normalizing flows directly on $\mathrm{SO}(3)$, showing promise for synthetic benchmarks but not extending to real-world robotics.  

In summary, probabilistic methods capture pose ambiguity and uncertainty better than deterministic ones but often trade efficiency or generality for expressiveness, motivating scalable methods that connect theoretical advances with robotic deployment.  

\subsection{Flows on Manifolds}
Flow matching~\cite{lipmanFlowMatchingGenerative2023} has emerged as an alternative to diffusion for generative learning. It trains continuous normalizing flows (CNFs) by regressing vector fields along probability paths, yielding simulation-free training, closed-form objectives, and faster inference. Chen and Lipman~\cite{chenFlowMatchingGeneral2024} extended this to Riemannian manifolds via \emph{Riemannian Flow Matching}, which generalizes conditional flow matching using geodesic or spectral premetrics.  

Several works apply manifold-aware flow matching across domains. In robotics, Braun et al.~\cite{braunRiemannianFlowMatching2024} introduced Riemannian Flow Matching Policies for efficient motion generation, while Funk et al.~\cite{funkActionFlowEquivariantAccurate2024} and Zhang and Gienger~\cite{zhangAffordancebasedRobotManipulation2025} demonstrated SE(3)-equivariant flows for action and affordance learning. Beyond robotics, Miller et al.~\cite{millerFlowMMGeneratingMaterials2024} used it for crystalline material discovery, and SE(3)-flow matching has been applied to protein backbone generation~\cite{bose2023se,huguet2024sequence}.  

These studies show the versatility of manifold-aware flows in robotics, materials science, and biology. However, applications on $\mathrm{SO}(3)$ or $\mathrm{SE}(3)$ remain largely confined to synthetic or simulation settings~\cite{hsiaoConfrontingAmbiguity6D,liuDelvingDiscreteNormalizing2023}. To our knowledge, our work is the first to employ flow matching on $\mathrm{SE}(3)$ for \emph{real-world 6D object pose estimation}, connecting manifold generative modeling with practical robotic deployment.

\section{Method}
Given an RGB-D input, our goal is to provide a sample-based estimate of the 6D object pose distribution rather than a single deterministic solution. 
Objects are localized using off-the-shelf detectors such as Mask R-CNN~\cite{he2017mask} or CNOS~\cite{nguyen2023cnos}, from which object-centric RGB crops and partial point clouds are extracted. 
These observations are encoded by geometric and visual encoders and fused with $\text{DiT}^{\star}$  blocks to drive conditional flow matching on the $\mathrm{SE}(3)$ manifold (Sec.~\ref{sec:pipeline}). 
This formulation enables efficient training, probabilistic sampling of multi-modal pose hypotheses, and naturally extends to pose tracking (Sec.~\ref{sec:flow}). 
For pose selection (Sec.~\ref{sec:poseselection}), we introduce two complementary strategies: a \textit{model-free clustering approach} that aggregates the sample-based hypotheses into consensus modes, and a \textit{model-based geometric scoring} that ranks hypotheses by their agreement with the 3D object model.
Finally, we show in Sec.~\ref{sec:tasks} how the learned $\mathrm{SE}(3)$ distributions enable active perception and uncertainty-aware downstream tasks, in particular \textit{grasp planning under ambiguity}, leading to safer and more reliable robotic manipulation.

\begin{figure*}[t]
    \centering
    \includegraphics[width=\linewidth]{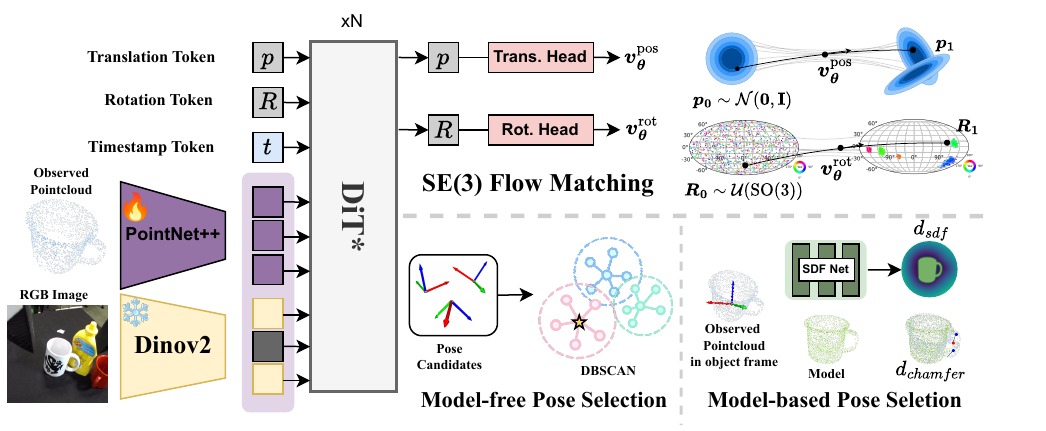}
    \caption{Overview of SE(3)-PoseFlow. Given an RGB-D input, we extract object-centric RGB crops and partial point clouds using off-the-shelf detectors. The visual and geometric features, together with timestep and sampled poses, are encoded and fused via $\text{DiT}^{\star}$ blocks with masked cross-attention to predict conditional velocity fields for SE(3) Flow Matching. The framework enables probabilistic sampling of multi-modal pose hypotheses and supports two complementary pose selection strategies: a model-free clustering approach and a model-based geometric scoring.}
    \label{fig:pipeline}
    \vspace{-15pt}
\end{figure*}

\subsection{Overall Pipeline}
\label{sec:pipeline}
Our framework integrates a dual-stream encoder for visual and geometric features, $\text{DiT}^{\star}$ blocks with masked cross-attention, and an $\mathrm{SE}(3)$ flow matching module (see Fig.~\ref{fig:pipeline}). 
The image stream employs a pretrained DINOv2 ViT~\cite{oquabDINOv2LearningRobust2024} to extract semantic patch embeddings, with the backbone kept frozen during training. 
In parallel, the point cloud stream uses a PointNet++~\cite{qi2017pointnet++} encoder, trained from scratch to preserve fine-grained spatial structure. 
To stabilize training, we first normalize partial point clouds by shifting them to zero-mean, which removes dependence on absolute camera-frame translations and instead emphasizes relative offsets in the object frame, thereby avoiding training collapse.
Both modalities are projected into a shared 256-dimensional feature space. 
The continuous flow timestamp $t$ is encoded via Fourier features and injected through adaptive layer normalization to condition the attention blocks.  

In place of the original DiT blocks~\cite{peeblesScalableDiffusionModels2023}, which employ self-attention for image-to-image generation, we introduce \emph{masked cross-attention blocks}. 
Separate pose tokens are assigned for translation in $\mathbb{R}^3$ and rotation in $\mathrm{SO}(3)$, thereby disentangling the two distributions. 
The cross-attention mechanism allows pose tokens to learn relations with image and point cloud features. 
A segmentation-derived binary mask specifies the set of \emph{active tokens} that participate in attention, filtering out background noise. 
Moreover, cross-attention reduces the computational complexity to linear $O(n)$ rather than quadratic $O(n^2)$.  

The refined pose tokens are decoded by translation and rotation heads to yield velocity fields $\boldsymbol{v_\theta^{\text{pos}}}$ and $\boldsymbol{v_\theta^{\text{rot}}}$. 
These velocity fields parameterize the conditional $\mathrm{SE}(3)$ flow for distribution sampling and integration.  

\subsection{Flow Matching on $\mathrm{SE}(3)$}
\label{sec:flow}



We model the conditional pose distribution $p(R,p \mid O,I)$, where $R \in \mathrm{SO}(3)$ and $p \in \mathbb{R}^3$.  
Here, $O$ denotes the observed point cloud and $I$ the corresponding image.  
Following the flow-matching framework~\cite{lipmanFlowMatchingGenerative2023,chenFlowMatchingGeneral2024}, we employ the \emph{Rectified Linear Flow (RLF)}, which defines a probability path between a random initialization $(R_0,p_0)$ and a target pose $(R_1,p_1)$.  
The translation is interpolated linearly in $\mathbb{R}^3$, while the rotation follows the geodesic on $\mathrm{SO}(3)$:
\begin{equation}
    p_t = (1 - t)\, p_0 + t \, p_1, 
    \qquad
    R_t = R_0 \exp\!\big( t \cdot \log(R_0^\top R_1) \big),
\end{equation}
for $t \in [0,1]$.  

Differentiating the path yields the ground-truth conditional velocity fields for the translation $\dot{p}_t \in \mathbb{R}^3$ and rotation $\dot{r}_t \in \mathbb{R}^3$
\begin{equation}
    \dot{p}_t = \tfrac{p_1 - p_t}{1 - t}, 
    \qquad 
    \dot{r}_t = \tfrac{1}{1 - t}\log(R_t^\top R_1).
\end{equation}
Here, $\log(\cdot)$ and $\exp(\cdot)$ denote the Lie algebra logarithm and exponential maps on $\mathrm{SO}(3)$.  

During training, the initial pose $(R_0,p_0)$ is sampled uniformly at random from $\mathrm{SE}(3)$, and the network predicts the conditional velocity fields
\[
\ell_{\text{pos}} = \big\| v_\theta^{\text{pos}}(p_t \mid O,I,t) - \dot{p}_t \big\|^2, 
\hspace{0.35em} 
\ell_{\text{rot}} = \big\| v_\theta^{\text{rot}}(R_t \mid O,I,t) - \dot{r}_t \big\|^2.
\]
The overall flow-matching objective is
\begin{equation}
    \mathcal{L}_{\text{FM}}(\theta) 
    = \mathbb{E}_{t,(R_0,p_0),(R_1,p_1)} 
      \big[ \lambda \, \ell_{\text{pos}} + \ell_{\text{rot}} \big],
\end{equation}
where $\lambda$ is a weighting factor between translation and rotation losses, set to $\lambda=10$ in our experiments to account for their different sensitivity scales.  

At inference time, pose samples are generated by starting with a set of randomly sampled pose candidates and integrating the learned vector field from $(R_0,p_0)$ to $t=1$ for each candidate. This process produces a set of pose hypotheses, which, depending on the degree of pose ambiguity in the observation, may either cluster around a single mode or exhibit greater diversity.  
For pose tracking, we naturally extend the framework by initializing $(R_0,p_0)$ from the previous estimate instead of starting from random samples, thereby enforcing temporal coherence across frames.

\subsection{Pose Selection}
\label{sec:poseselection}

We study two complementary strategies for selecting representative poses: a \textit{model-free} clustering approach in $\mathrm{SE}(3)$ and a \textit{model-based} geometric evaluation.  

\textbf{Model-free Pose Selection} 
Given a hypothesis set $\{T_i\}$, we apply DBSCAN~\cite{ester1996density} with a distance that combines rotational and translational differences:
\begin{equation}
d(T_1,T_2)=\sqrt{\left(\tfrac{\theta(R_1,R_2)}{\epsilon_R}\right)^2+\left(\tfrac{\|t_1-t_2\|}{\epsilon_t}\right)^2},
\end{equation}
where $\theta(\cdot,\cdot)$ denotes the geodesic angle between two rotations.  
The largest cluster is selected and its representative is computed as the Karcher mean~\cite{moakher2002means} on $\mathfrak{se}(3)$.  
This approach requires no object model and adapts naturally to multi-modal distributions.  
In practice we use $\epsilon_R=10^\circ$ and $\epsilon_t=0.03$cm.  

\textbf{Model-based Pose Selection} 
To further resolve ambiguity, we evaluate pose candidates against the object geometry using two objectives.  
The \emph{Chamfer loss}~\cite{fan2017point} measures the one-sided distance from observed points $P$ to the transformed model points $M$: 
\begin{equation}
d_{\text{chamfer}}(P,M)=\tfrac{1}{|P|}\sum_{p\in P}\min_{m\in T(M)}\|p-m\|_2^2 .
\end{equation}
It is simple and requires no pre-training, but is sensitive to point density and observation noise.  
The \emph{SDF loss}~\cite{park2019deepsdf,gropp2020igr,yariv2021volsdf,wang2021neus} employs a learned signed distance function $f$, with coordinates normalized to $[-1,1]^3$, to compute
\begin{equation}
d_{\text{sdf}}(P,T)=\tfrac{1}{|P|}\sum_{p\in P} f\!\left(\text{Norm}(T^{-1}p)\right)^2 .
\end{equation}
This provides continuous and global surface feedback, but requires high-quality meshes or dense point clouds for training, and may fail with sparse supervision.  

For both objectives, we convert residuals into log-likelihood scores and retain the top 20\% of poses.  
Chamfer loss thus serves as a lightweight, training-free baseline, while SDF offers stronger geometric consistency when reliable shape supervision is available.

\subsection{Exploiting the Sample-based Pose Estimation for Downstream Tasks}
\label{sec:tasks}

\textbf{Active Perception}~~~In realistic scenarios, the robot's sensors often provide only partial observations due to occlusions or restricted viewpoints. 
To reduce pose uncertainty, we leverage the covariance of pose hypotheses. 
Given a set of sampled transformations $\{T_i\}$, we estimate the mean and covariance of translations and rotations, where rotational uncertainty is quantified in the tangent space of $\mathrm{SO}(3)$. 
Based on this uncertainty, the robot actively selects the next-best viewpoint on an object-centric viewing sphere at a fixed distance. 
Formally, the next viewpoint $v^\ast$ is chosen to maximally reduce the predicted rotational covariance:
\begin{equation}
    v^\ast = \arg\min_{v \in \mathcal{V}} 
        \; \mathbb{E}\!\left[ \mathrm{tr}\big(\Sigma_{R} \mid v \big) \right],
\end{equation}
where $\mathcal{V}$ denotes the discrete set of admissible viewpoints, and $\Sigma_{R}$ is the covariance of rotations estimated from $\{T_i\}$. 
Since this objective has no closed-form solution, we approximate it by particle sampling: candidate viewpoints are uniformly sampled on the viewing sphere, their induced pose covariances are evaluated, and the viewpoint with minimal rotational uncertainty is selected as the next best view. This strategy allows the robot to actively move its sensor around the object to disambiguate symmetric configurations and acquire confident pose estimates with minimal exploration cost.

\textbf{Robotic Grasping}
In cluttered scenes, grasp planning must explicitly account for uncertainty in object orientation. We synthesize grasps by marginalizing over the pose candidates at the \emph{velocity} level. Let \(M=\{x_i\}_{i=1}^N\subset\mathbb{R}^3\) denote the object point cloud and \(\{T_k\}_{k=1}^K\subset\mathrm{SE}(3)\) be pose hypotheses drawn (e.g. by sampling or clustering) from the posterior \(p(T\mid O,I)\).
For each hypothesis, we transform the model into the world frame and then normalize the transformed cloud into a canonical space:
\[
M_k^{(w)}=\{T_k x_i\}_{i=1}^N,\qquad
\tilde M_k \;=\; M_k^{(w)} \;-\; \frac{1}{N}\sum_{i=1}^N M_{k,i}^{(w)}.
\]
Given the models representing the pose hypothesis, we then leverage a flow-matching-based generative model, i.e., EquiGraspFlow~\cite{lim2024equigraspflow}, for grasp synthesis.
In particular, starting from randomly initialized grasp pose candidates, we obtain the velocity update vectors for every combination of grasp pose and pose hypothesis (represented by the different models) and form the pose (hypothesis) marginal (mean) velocity field:
\begin{equation}
\bar v(M,t)\;\approx\;\frac{1}{K}\sum_{k=1}^{K} v_\theta\!\big(\tilde M_k,t\big).
\label{eq:meanfield_canonical_final}
\end{equation}
Integrating \(\bar v\) yields grasp samples that are consistent with the multiple pose hypotheses and thereby clustered across high-probability pose modes.
For instance, for a mug with an occluded handle (or ambiguous azimuth), averaging velocities on the canonical space suppresses side-grasp modes tied to uncertain yaw and concentrates mass on top grasps that remain valid across the plausible orientations.

\begin{figure}[t]
    \centering
    \includegraphics[width=.95\linewidth]{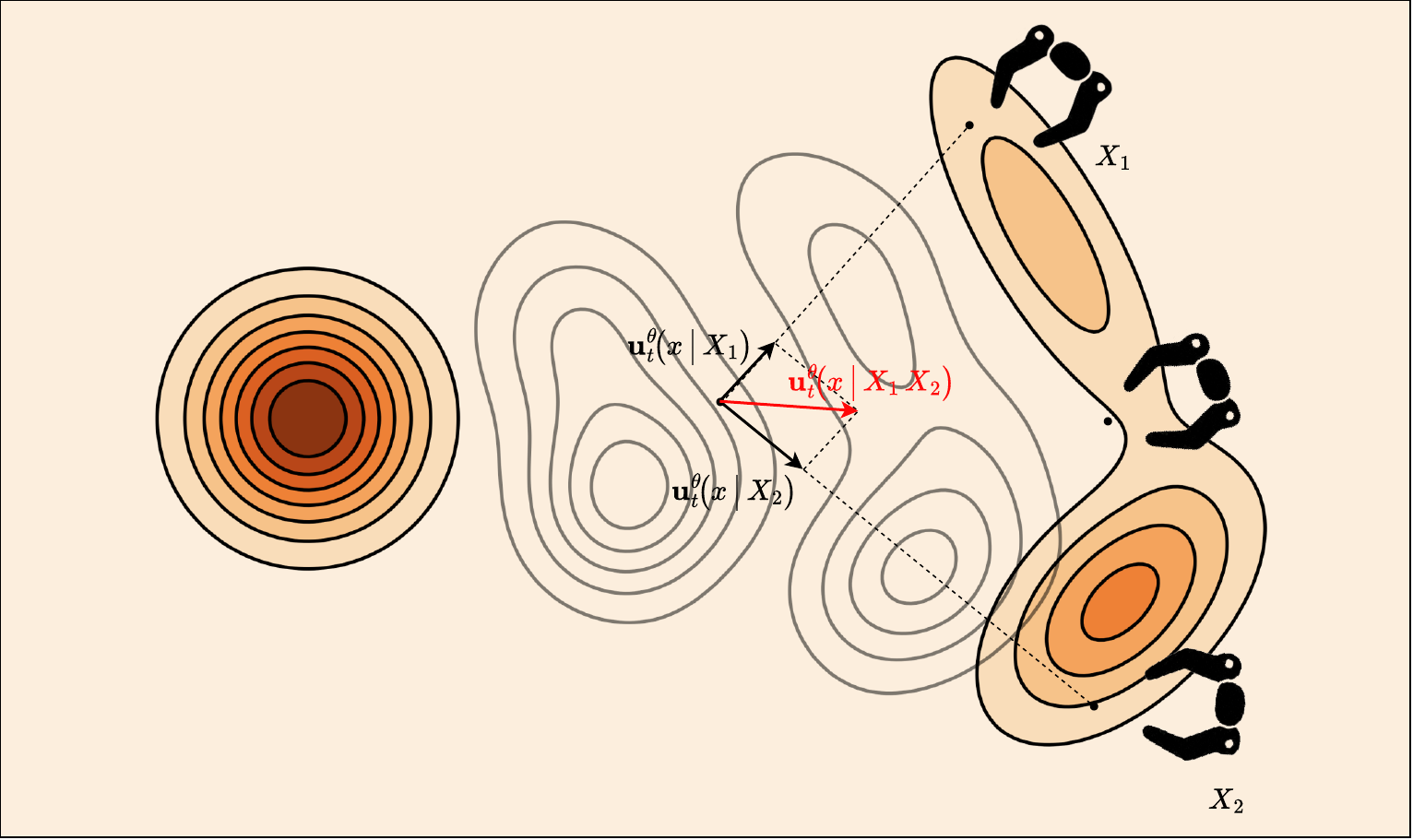}
\caption{\textbf{Illustrating the mean grasp pose velocity under pose uncertainty.}
EquiGraspFlow velocities are averaged per pose hypothesis to form a mean field, which is integrated to sample grasps that are robust to pose ambiguity (e.g., favouring top grasps for a mug with an occluded handle).}
    \label{fig:grasp_flow}
\end{figure}




\section{Experiments}

\textbf{Dataset} We conduct experiments on three widely used benchmarks for 6D object pose estimation.
REAL275~\cite{wang2019normalized} contains real RGB-D sequences of 6 object categories with large intra-class variations, which serves as the standard benchmark for evaluating category-level methods. 
YCB-Video (YCB-V)~\cite{xiang2017posecnn} consists of 92 video sequences of 21 YCB objects in cluttered scenes, widely adopted for instance-level evaluation. 
LINEMOD-Occlusion (LM-O)~\cite{hinterstoisser2012model} contains 8 textureless household objects under heavy occlusion, and is commonly used to benchmark instance-level methods in challenging real-world scenes. 
Together, these three datasets provide a comprehensive evaluation setting, covering both category-level generalization and challenging instance-level scenarios. 

\textbf{Evaluation Metrics} Following GenPose~\cite{zhangGenPoseGenerativeCategorylevel2023} and DiffusionNOCS~\cite{ikeda2024diffusionnocs}, we evaluate pose accuracy using rotation and translation thresholds.
A prediction is considered correct if its rotation error is below $\alpha$ degrees and its translation error is below $\beta$ centimeters. 
We report results under the commonly used $5^\circ\!2$cm, $5^\circ\!5$cm, and $10^\circ\!5$cm criteria, averaged across all objects and scenes. 
For symmetric objects, the minimum geodesic rotation error over the discrete symmetry set is adopted.  

\textbf{Baseline} We compare against both deterministic and probabilistic approaches. 
Deterministic baselines include NOCS~\cite{wang2019normalized}, DualPoseNet~\cite{lin2021dualposenet} and SPD~\cite{tian2020shape}, which are representative methods reported on the NOCS benchmark. 
For these methods, we report the results listed on the official leaderboard of~\cite{zhangGenPoseGenerativeCategorylevel2023}. 
Probabilistic baselines include GenPose~\cite{zhangGenPoseGenerativeCategorylevel2023} and DiffusionNOCS~\cite{ikeda2024diffusionnocs}. 
We directly evaluate them using their publicly available implementations and pretrained checkpoints. 
Since DiffusionNOCS does not release code for recovering poses from NOCS maps, we follow the protocol described in~\cite{wang2019normalized} to estimate object poses from the predicted NOCS maps.
For fair comparison, all methods are evaluated on the same test splits under the unified protocol without per-object tuning.

\begin{figure}[ht]
\centering
\includegraphics[width=\linewidth]{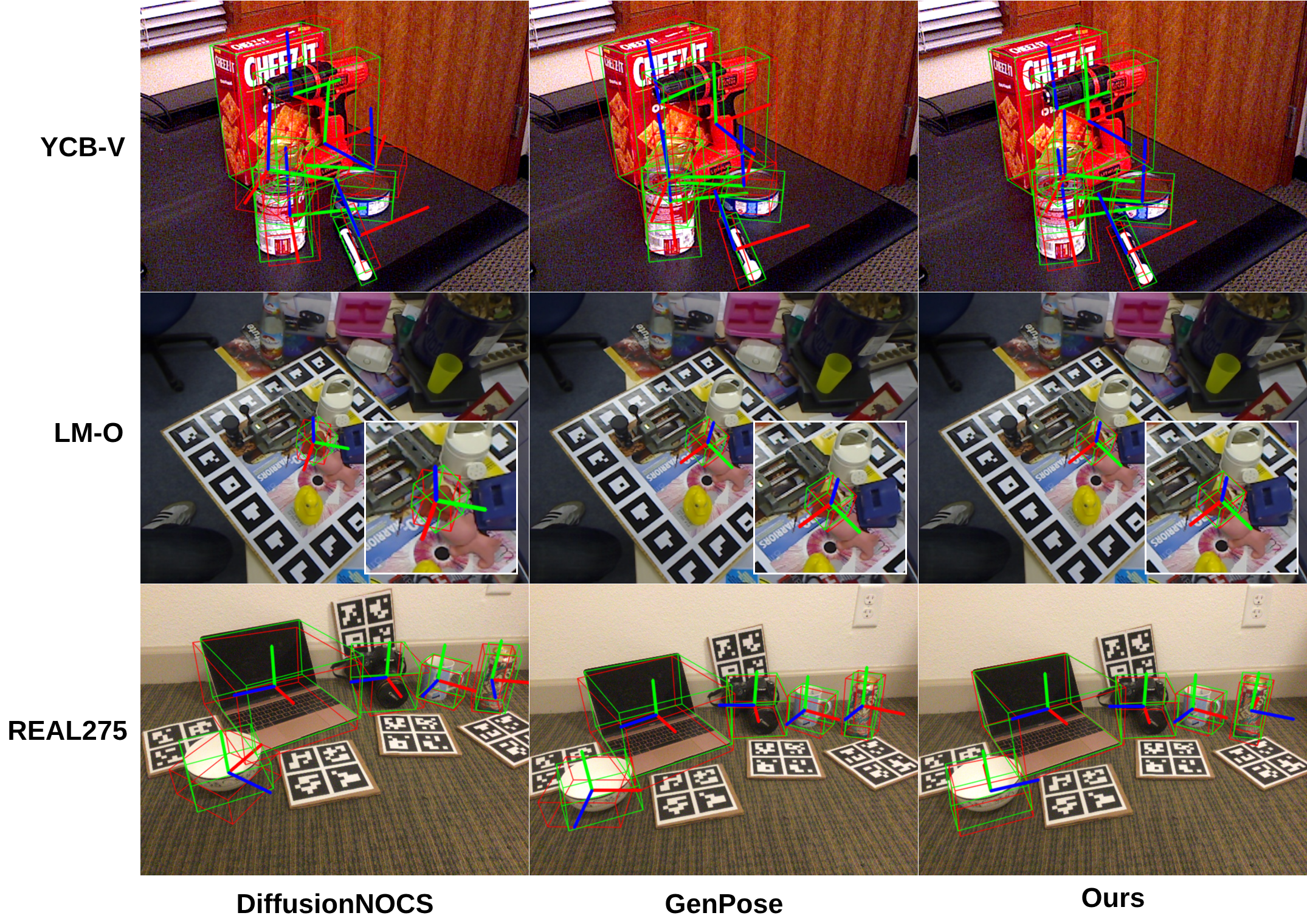}
\caption{Qualitative comparison of pose estimation on  YCB-V, LM-O and Real275 datasets.}
\label{fig:quan_results}
\end{figure}

\subsection{Quantitative Analysis}
\textbf{Results on Real275} 
Following the evaluation protocol of GenPose~\cite{zhangGenPoseGenerativeCategorylevel2023}, 
we sample 50 pose hypotheses per object and retain the top 40\% according to the pose selection strategy. 
Since Real275 does not provide meshes or dense point clouds, SDF-based scoring cannot be applied and Chamfer distance is used for re-ranking. 
Table~\ref{tab:real275_results} shows that probabilistic methods consistently outperform regression-based baselines such as NOCS, DualPoseNet, and SPD, underlining the benefit of explicitly modeling multiple pose hypotheses in ambiguous settings. 
Relative to DiffusionNOCS, our approach achieves higher accuracy across most thresholds. 
This improvement stems from differences in inference design: DiffusionNOCS predicts normalized object coordinate (NOC) maps from masked RGB input and subsequently computes poses in SE(3), a process that is sensitive to depth--NOC misalignment and mask errors. 
Our method directly samples and evaluates hypotheses in SE(3), ensuring geometric consistency and yielding more reliable estimates under sparse or noisy depth. 

At the strictest thresholds ($5^\circ$2cm and $5^\circ$5cm), our accuracy is slightly lower than GenPose. 
This is mainly due to the scoring stage: Chamfer distance is sensitive to the sparse and noisy point clouds provided in Real275, reducing its effectiveness. 
GenPose instead employs a learned energy network to approximate dataset-specific likelihoods, which is effective on Real275 but less transferable across datasets. 

\textbf{Results on BOP Datasets} 
The BOP benchmarks (YCB-V and LM-O) provide denser RGB-D input and high-quality mesh supervision, which allows our probabilistic pipeline to fully exploit geometry-aware scoring (Table~\ref{tab:bop_results}). 
SDF-based re-ranking provides a smoother and more global error landscape than Chamfer distance, improving the separation between valid poses and structurally inconsistent hypotheses. 
With this strategy, our method achieves the highest accuracy on LM-O and competitive results on YCB-V, outperforming prior probabilistic approaches. 
These findings support our hypothesis that probabilistic sampling combined with geometry-aware re-ranking improves the handling of pose ambiguity, particularly when strong geometric supervision is available. Qualitative results visualizing pose estimates on Real275, YCB-V and LM-O are shown in Fig.~\ref{fig:quan_results}.

\begin{table}[t]
\centering
\vspace{10pt}
\caption{
Quantitative comparison of category-level object pose estimation on REAL275 dataset.}
\vspace{5pt}
\resizebox{\columnwidth}{!}{
\begin{tabular}{l|l|cccc}
\toprule
\multicolumn{2}{c|}{Method} & $5^{\circ}2$cm$\uparrow$ & $5^{\circ}5$cm$\uparrow$ & $10^{\circ}2$cm$\uparrow$ & $10^{\circ}5$cm$\uparrow$ \\ 
\midrule
\multirow{3}{*}{Deterministic} 
& NOCS~\cite{wang2019normalized}        & -    & 9.5  & 13.8  & 26.7  \\
& DualPoseNet~\cite{lin2021dualposenet} & 29.3 & 35.9 & 50.0  & 66.8  \\
& SPD~\cite{tian2020shape}              & 19.3 & 21.4 & 43.2  & 54.1  \\
\midrule
\multirow{3}{*}{Probabilistic}
& DiffusionNOCS~\cite{ikeda2024diffusionnocs}       & -    & 35.0 & 66.6 & 77.1  \\
& GenPose~\cite{zhangGenPoseGenerativeCategorylevel2023} & \textbf{52.1} & \textbf{60.9} & 72.4 & 84.0 \\
\rowcolor{gray!15}
& Ours                  & 48.8    &   56.3   &  \textbf{76.3}   & \textbf{89.1}   \\
\bottomrule

\end{tabular}
}
\vspace{-10pt}
\label{tab:real275_results}
\end{table}

\begin{table}[t]
\centering
\vspace{10pt}
\caption{
Comparison of generative model-based 6D object pose estimation methods on the BOP dataset.}
\resizebox{\columnwidth}{!}{
\begin{tabular}{l|cc|cc}
\toprule
\multirow{2}{*}{Method} 
& \multicolumn{2}{c|}{YCB-V} 
& \multicolumn{2}{c}{LM-O} \\
& $5^{\circ}5$cm$\uparrow$ & $10^{\circ}5$cm$\uparrow$ 
& $5^{\circ}5$cm$\uparrow$ & $10^{\circ}5$cm$\uparrow$ \\
\midrule
DiffusionNOCS~\cite{ikeda2024diffusionnocs}                   & 23.4 & 54.8 & 15.5 & 42.5 \\
GenPose~\cite{zhangGenPoseGenerativeCategorylevel2023}        & \textbf{46.2} & 63.8 & 32.2 & 48.2 \\
\rowcolor{gray!15}
Ours 
 & 45.4 & \textbf{68.2} & \textbf{35.2} & \textbf{53.7} \\
\bottomrule
\end{tabular}
}
\label{tab:bop_results}
\end{table}

\begin{table}[ht]
\centering
\resizebox{\linewidth}{!}{%
\begin{tabular}{lcccccc}
\toprule
& \multicolumn{2}{c}{REAL275} & \multicolumn{2}{c}{YCB-V} & \multicolumn{2}{c}{LM-O} \\
\cmidrule(lr){2-3} \cmidrule(lr){4-5} \cmidrule(lr){6-7}
& $5^{\circ}5\text{cm} \uparrow$ & $10^{\circ}5\text{cm} \uparrow$
& $5^{\circ}5\text{cm} \uparrow$ & $10^{\circ}5\text{cm} \uparrow$
& $5^{\circ}5\text{cm} \uparrow$ & $10^{\circ}5\text{cm} \uparrow$ \\
\midrule
w/o RGB     &  \textbf{52.3} & 77.9  & 20.9 & 43.6  & \textbf{30.5}  &  46.7 \\
w/o kv mask & 47.8  & 73.8  & 36.4 & 51.2 & 22.8  & 40.2 \\
\midrule
ours        & 51.3  & \textbf{79.3} & \textbf{40.2}  & \textbf{54.7}& 30.1 & \textbf{48.3}  \\
\bottomrule
\end{tabular}}
\caption{Ablation on input modalities and kv masking under two thresholds.}
\label{tab:design_choices}
\end{table}

\subsection{Ablation Study}
\label{sect:AblationStudy}


\textbf{Input modalities and attention mask}
We conduct an ablation study to examine the influence of input modalities and the proposed masking mechanism in Sec.~\ref{sec:pipeline}. 
To remove stochasticity, all experiments are performed without pose selection. 
Table~\ref{tab:design_choices} compares three settings: \textit{W/o RGB}, which employs only point clouds as input; \textit{W/o kv mask}, which removes the masking applied to visual tokens; and \textit{Ours}, which integrates both visual and geometric features with the mask-attention design.  

On REAL275, using only point clouds slightly outperforms the RGB-augmented variants. 
This suggests that visual features are less useful when occlusions are limited and objects are mostly symmetric and texture-less, as is the case in REAL275 where only the camera and laptop categories contain significant textures. 
On LM-O, performance differences remain marginal for similar reasons. 
In contrast, on YCB-V, which contains many textured objects, the addition of RGB improves accuracy by more than 10\%, highlighting the importance of visual cues in textured scenes. 
Across all datasets, models with kv mask consistently perform better than those without, indicating that the mask helps suppress background clutter and extract cleaner, more generalizable visual features.

\begin{table}[ht]
\centering
\resizebox{\linewidth}{!}{%
\begin{tabular}{lcccc}
\toprule
& \multicolumn{2}{c}{Real275} & \multicolumn{2}{c}{YCB-V} \\
\cmidrule(lr){2-3} \cmidrule(lr){4-5}
 & $5^\circ$5cm $\uparrow$ & $10^\circ$5cm $\uparrow$
 & $5^\circ$5cm $\uparrow$ & $10^\circ$5cm $\uparrow$ \\
\midrule
None                    &  51.3  &  79.3  &  40.2  &  54.7  \\
Model-free              &  52.5  &  88.2  &  42.5  &  65.6  \\
Model-based (Chamfer)   &  56.3  &  89.1  &  43.1  &  67.5  \\
Model-based (SDF)       &  -  &  -  &  \textbf{45.4}  &  \textbf{68.2}  \\
\bottomrule
\end{tabular}}
\caption{Ablation study on pose selection strategies under two accuracy thresholds on Real275 and YCB-V.}
\label{tab:pose_selection}
\end{table}

\textbf{Pose Selection} 
Probabilistic sampling inevitably generates outlier hypotheses, making robust pose selection essential for reliable performance.
We compare four strategies under a unified setting of 50 samples in Table~\ref{tab:pose_selection}: 
(i) \emph{None}, where poses are taken without selection, 
(ii) \emph{Model-free}, which evaluates all candidates in the best cluster, 
(iii) \emph{Model-based (Chamfer)}, ranking poses by Chamfer distance to the object model, and 
(iv) \emph{Model-based (SDF)}, which scores poses using a neural signed distance function. 
For model-based methods, we retain the top $40\%$ hypotheses, while the model-free variant evaluates all cluster members.  

Model-free selection improves over None by filtering spurious outliers, but remains weaker than model-based approaches due to the lack of geometric priors. 
On Real275, no mesh or dense point cloud is available, causing SDF training to collapse; hence results are missing. 
On YCB-V, SDF-based scoring achieves the best accuracy, confirming its advantage over Chamfer distance in handling small alignment errors. 
Overall, these results indicate that model-based scoring, particularly with SDF supervision, is key to fully exploiting probabilistic sampling for resolving pose ambiguity. 


\begin{table}[ht]
\centering
\resizebox{\linewidth}{!}{%
\begin{tabular}{lcccccc}
\toprule
 & \multicolumn{2}{c}{Pose Estimation} & \multicolumn{2}{c}{Pose Tracking} & Runtime \\
\cmidrule(lr){2-3} \cmidrule(lr){4-5} \cmidrule(lr){6-6}
Steps & 5$^\circ$5cm $\uparrow$ & 10$^\circ$5cm $\uparrow$ 
      & 5$^\circ$5cm $\uparrow$ & 10$^\circ$5cm $\uparrow$ 
      & Speed(FPS) $\uparrow$ \\
\midrule
1   & 7.6  & 20.6 & 51.2 & 83.6 & \textbf{35.2} \\
2   & 44.4 & 75.7 & \textbf{57.4} & 88.2 & 21.3 \\
3   & 52.5 & 81.3 &  56.2  & 86.3  & 16.7 \\
5   & 56.3 & \textbf{89.1} &  56.9  &  87.1 & 10.1 \\
10  & \textbf{57.8} & 88.6 &  57.2  &  \textbf{88.8}  &  4.3 \\
\bottomrule
\end{tabular}}
\caption{Ablation study on the effect of varying the number of available inference steps for pose estimation and tracking on REAL275.}
\label{tab:steps_ablation}
\end{table}

\textbf{Inference Steps across Pose Estimation and Tracking} 
We study the effect of varying the number of ODE integration steps on the pose estimation and tracking quality on REAL275 in Table~\ref{tab:steps_ablation}. 
For pose estimation, we follow the model-based protocol with 50 pose samples per frame. 
For tracking, the first frame is initialized from a perturbed ground truth pose (rotation up to $20^\circ$, translation up to 5 cm), and subsequent frames use the previous prediction.  

Increasing the number of inference steps improves the pose estimation results for up to 5 steps, after which the accuracy saturates while the runtime increases. 
Tracking, however, achieves strong performance even with a single inference step, as the model only needs to refine a near-correct initialization. 
These results demonstrate the efficiency of the rectified flow matching objective: accurate results can be obtained within very few steps, owing to its continuous and constant velocity field formulation. 

\subsection{Uncertainty-aware Robotic Tasks}
\textbf{Active Perception} 
We validate our active perception strategy on a Franka Panda arm equipped with a ZED Mini RGB-D camera mounted on the wrist. 
To obtain consistent object masks under varying viewpoints, we employ SAM2~\cite{raviSAM2Segment} for mask tracking and extend it with dynamic prompts in multi-object scenes. 
Viewpoints are sampled on an object-centric sphere at a fixed radius, orienting the camera toward the object center. 
Each candidate viewpoint is scored by the induced pose covariance, and the next-best view is selected as the one minimizing rotational uncertainty, as illustrated in Fig.~\ref{fig:active_perception}. 
The robot subsequently executes the chosen motion, actively moving the camera around the object to reduce ambiguity and improve the pose estimate. 
This setup enables quantitative evaluation of uncertainty reduction and pose accuracy in real-world conditions. 

\begin{figure}[ht]
\centering
\includegraphics[width=\linewidth]{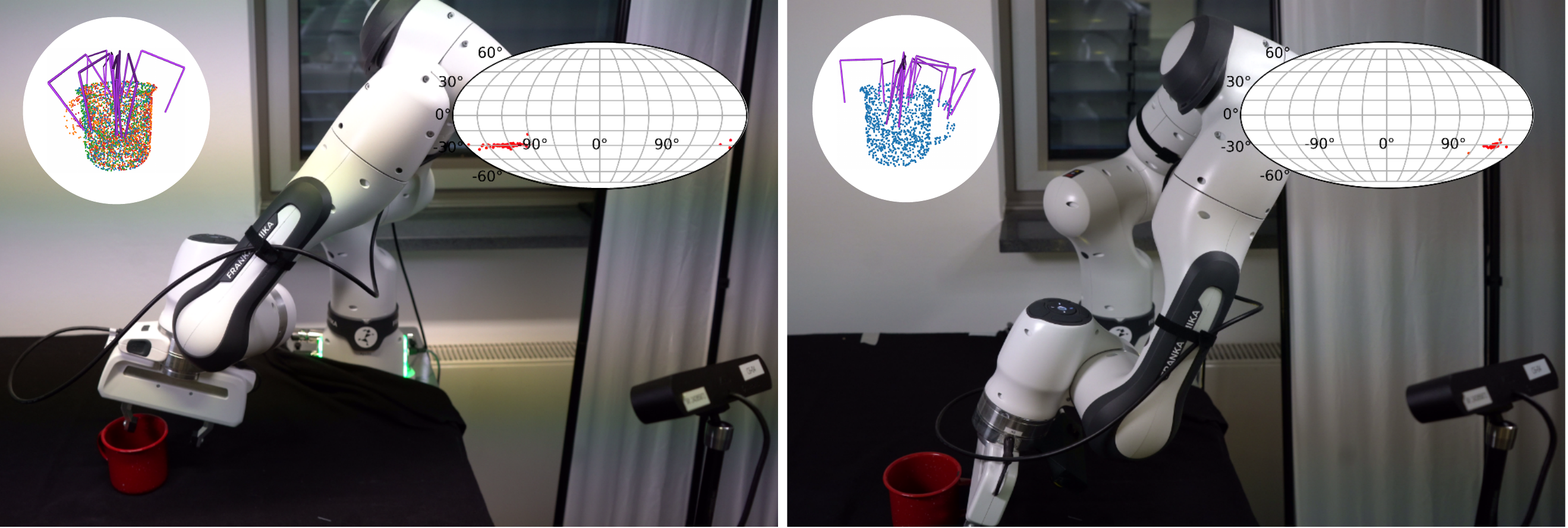}
\caption{Uncertainty-aware grasping on a mug. 
\textbf{Left:} Occluded case with a multi-modal sample-based distribution of pose hypotheses; Sampling grasps using EquiGraspFlow while marginalizing over the multiple pose hypotheses generates top-down grasps that remain valid across all pose hypotheses. 
\textbf{Right:} Non-occluded case with a unimodal distribution, i.e., all the samples agree on a single pose hypothesis; Sampling grasps using EquiGraspFlow while marginalizing over the multiple pose hypotheses (which now coincide to one pose) also produces side grasps targeting the handle.}
\label{fig:grasp_exp}
\end{figure}
\textbf{Uncertainty-Aware Grasping Task.}
The grasping experiments are conducted using an external ZED2 RGB-D camera and a Franka Panda arm controlled through ROS/MoveIt. 
Grasp poses are generated with EquiGraspFlow~\cite{lim2024equigraspflow} as described in Sec.~\ref{sec:tasks}: the observed point cloud is transformed into the canonical frame, the mean velocity field is computed, and 10 candidate grasps are sampled. 
The nearest feasible grasp is then executed on the robot. 
We evaluate two scenarios with a mug: one where the handle is clearly visible (non-occluded) and another where the handle is self-occluded. 
In the non-occluded case, the pose distribution is unimodal, and the grasp pose generation therefore also yields side grasps reaching for the handle. 
In contrast, under self-occlusion the pose distribution becomes multi-modal due to the ambiguous yaw angle, and the proposed grasp sampling automatically shifts to top-down grasps that remain valid across all pose hypotheses. 
As a baseline, we consider grasp generation based on only a single pose hypothesis.
While this works in the non-occluded case, it fails under occlusion as it continues to propose handle-reaching side grasps. 
Our uncertainty-aware approach, by contrast, consistently selects top-down grasps that succeed across all plausible poses. 
Quantitatively, we executed 10 trials per visibility condition on the same mug. 
The results in Table~\ref{tab:mug_grasp_eval} show that the baseline variant, \emph{EquiGraspFlow (single)}, achieves a success rate of \textbf{75.0\%} overall (9/10 in the non-occluded case and 6/10 under occlusion), whereas our uncertainty-aware \emph{EquiGraspFlow (multi)} reaches \textbf{95.0\%}, performing perfectly when the handle is visible and remaining robust under self-occlusion. 
This confirms that marginalizing grasp sampling over multiple pose hypotheses substantially improves grasp reliability in the presence of occlusion and pose ambiguity. 
Qualitative examples are shown in Fig.~\ref{fig:grasp_exp}.

\begin{table}[ht]
\centering
\resizebox{\linewidth}{!}{%
\begin{tabular}{lccc}
\toprule
 & Non-occluded & Occluded & Total (\#/20) \\
\midrule
EquiGraspFlow (single) & 9/10 & 6/10 & 15/20 (75.0\%) \\
\textbf{EquiGraspFlow (multi)} & 10/10 & 9/10 & \textbf{19/20 (95.0\%)} \\
\bottomrule
\end{tabular}}
\caption{Real-robot evaluation of uncertainty-aware grasping on a mug.
Each cell reports 10 grasp attempts under two visibility conditions:
\textbf{Non-occluded} (handle visible, unimodal distribution) and
\textbf{Occluded} (handle self-occluded, multi-modal yaw ambiguity).
The baseline uses a single pose hypothesis, while ours marginalizes over multiple pose hypotheses.}
\label{tab:mug_grasp_eval}
\end{table}

\section{Conclusion}
We presented a probabilistic framework for 6D object pose estimation based on $\mathrm{SE}(3)$ flow matching. 
Unlike deterministic regressors, our method generates sample-based hypotheses that capture multi-modality and calibrated uncertainty, which is crucial for handling symmetries, occlusions, and partial observability. 
The integration of visual and geometric cues through DiT blocks with masked cross-attention enables robust performance across challenging benchmarks. 
We further showed that the resulting pose hypotheses can be directly exploited in downstream robotics tasks such as active perception and uncertainty-aware grasp generation.  

Our approach still has limitations. 
It does not yet generalize seamlessly to all object categories, and the modality gap between images and point clouds makes it difficult to obtain a unified representation—point maps may provide a promising alternative. 
Moreover, the framework is sample-based, and how to incorporate Bayesian inference for principled utilization of these samples remains an open question.  

Future work will extend this framework to multi-object scenes and long-horizon manipulation, where reasoning about joint pose distributions and temporal consistency is critical, and will explore sequential Bayesian methods such as particle filtering for robust online tracking.


\bibliographystyle{IEEEtran}
\bibliography{bibliography/reference}

\end{document}